\documentclass[11pt,a4paper]{article}
\usepackage[hyperref]{emnlp-ijcnlp-2019}
\usepackage{latexsym}
\usepackage{times,subcaption}
\usepackage{graphicx}
\usepackage{url}
\usepackage{booktabs}
\usepackage{amsmath,amsthm,amssymb,pstricks,pst-node}
\usepackage{xspace}
\usepackage{color}
\usepackage{soul}

\newsavebox\FrameBox
\newenvironment{Frame}{%
   \par\setbox\FrameBox\hbox\bgroup\minipage{0.45\textwidth}\parskip\baselineskip\ignorespaces
}{%
   \endminipage\egroup\fbox{\box\FrameBox}\par
}

\DeclareMathOperator*{\argmax}{arg\,max}
\aclfinalcopy 

\newcommand{\elmo}{ELMo\xspace}
\newcommand{\entelmo}{EntELMo\xspace}
\newcommand{\bert}{BERT\xspace}

\newcommand{\enteval}{EntEval\xspace}
\newcommand{\wikilink}{\textsc{WikiEnt}\xspace}
\newcommand{\fever}{FEVER\xspace}

\newcommand{\kore}{KORE\xspace}
\newcommand{\wikisrs}{WikiSRS\xspace}
\newcommand{\conceptnet}{ConceptNet\xspace}

\newcommand{\yago}{CoNLL-YAGO\xspace}
\newcommand{\contextrep}{contextualized entity representations\xspace}
\newcommand{\descrep}{descriptive entity representations\xspace}

\definecolor{dkgreen}{rgb}{0,0.6,0}

\newenvironment{itemizesquish}{\begin{list}{\labelitemi}{\setlength{\itemsep}{-0.2em}\setlength{\labelwidth}{0.5em}\setlength{\leftmargin}{\labelwidth}
\addtolength{\leftmargin}{\labelsep}}}{\end{list}}

\newenvironment{enumeratesquish}{\begin{list}{\addtocounter{enumi}{1}\labelenumi}{\setlength{\itemsep}{0em}\setlength{\labelwidth}{0.5em}\setlength{\leftmargin}{\labelwidth}\addtolength{\leftmargin}{\labelsep}}}{\end{list}\setcounter{enumi}{0}}

\title{\enteval: A Holistic Evaluation Benchmark for Entity Representations}

\author{Mingda Chen$^{3}$\thanks{~~Equal contribution. Listed in alphabetical order.} \quad Zewei Chu$^{1*}$\quad Yang Chen$^{2}$ \quad Karl Stratos$^{4}$\thanks{~~Work done while the author was at Toyota Technological Institute at Chicago.} \quad Kevin Gimpel$^{3}$ \\
$^{1}$University of Chicago, IL, USA\\
$^{2}$Ohio State University, OH, USA\\
$^{3}$Toyota Technological Institute at Chicago, IL, USA\\
$^{4}$Rutgers University, NJ, USA\\
  \small\texttt{\{mchen,kgimpel\}@ttic.edu,stratos@cs.rutgers.edu,zeweichu@uchicago.edu,chen.9279@osu.edu}
}

\date{}

\begin{document}
\maketitle
\begin{abstract}
Rich entity representations are useful for a wide class of problems involving entities. Despite their importance, there is no standardized benchmark that evaluates the overall quality of entity representations. In this work, we propose \enteval: a test suite of diverse tasks that require nontrivial understanding of entities including entity typing, entity similarity, entity relation prediction, and entity disambiguation. In addition, we develop training techniques for learning better entity representations by using natural hyperlink annotations in Wikipedia. We identify effective objectives for incorporating the contextual information in hyperlinks into state-of-the-art pretrained language models \citep{peters-etal-2018-deep} and show that they improve strong baselines on multiple \enteval tasks.\footnote{Data processing and evaluation scripts are available at
{\href{https://github.com/ZeweiChu/EntEval}{\nolinkurl{https://github.com/ZeweiChu/EntEval}}}
}

\end{abstract}

\section{Introduction}

Entity representations play a key role in numerous important problems including language modeling~\cite{ji-etal-2017-dynamic}, 
dialogue generation~\cite{he-etal-2017-learning}, entity linking~\cite{gupta-etal-2017-entity}, and story generation~\cite{clark-etal-2018-neural}. 
One successful line of work on learning entity representations has been learning \emph{static} embeddings: 
that is, assign a unique vector to each entity in the training data
\citep{gupta-etal-2017-entity, yamada-etal-2016-joint, yamada2017learning}. 
While these embeddings are useful in many applications, they have the obvious drawback of 
not accommodating unknown entities. 
Another limiting factor is the lack of an evaluation benchmark:
it is often difficult to know which entity representations are better for which tasks. 

We introduce \enteval: a carefully designed benchmark for holistically evaluating entity representations. 
It is a test suite of diverse tasks that require nontrivial understanding of entities,  
including entity typing, entity similarity, entity relation prediction, and entity disambiguation. Motivated by the recent success of contextualized word representations (henceforth: CWRs) from pretrained models~\citep{NIPS2017_7209,peters-etal-2018-deep,devlin2018bert,yang2019xlnet,liu2019roberta}, we propose to encode the mention context or 
the description to dynamically represent an entity. 
In addition, we perform an in-depth comparison of ELMo and BERT-based embeddings 
and find that they show different characteristics on different tasks. 
We analyze each layer of the CWRs and make the following observations: 
\begin{itemizesquish}
\item The dynamically encoded entity representations show a strong improvement on the entity disambiguation task compared to prior work using static entity embeddings.
\item BERT-based entity representations require further supervised training to perform well on downstream tasks, while ELMo-based representations are more capable of performing zero-shot tasks. 
\item In general, higher layers of ELMo and BERT-based CWRs are more transferable to \enteval tasks. 
\end{itemizesquish}

To further improve contextualized and descriptive entity representations (CER/DER), we leverage natural hyperlink annotations in Wikipedia. 
We identify effective objectives for incorporating the contextual information in hyperlinks 
and improve ELMo-based CWRs on a variety of entity related tasks. 

\section{Related Work}

\enteval and the training objectives considered in this work are built on previous works that involve reasoning over entities. We give a brief overview of relevant works. 

\paragraph{Entity linking/disambiguation.}

Entity linking is a fundamental task in information extraction 
with a wealth of literature~\citep{he-etal-2013-learning,Guo:2014:REL:2661829.2661887,ling-etal-2015-design, huang2015leveraging,francis-landau-etal-2016-capturing,le-titov-2018-improving,martins-etal-2019-joint}.
The goal of this task is to map a mention in context to the 
corresponding entity in a database. A natural approach is to learn 
entity representations that enable this mapping. 
Recent works focused on learning a fixed embedding for each entity 
using Wikipedia hyperlinks~\citep{yamada-etal-2016-joint,ganea-hofmann-2017-deep,le-titov-2019-boosting}. 
\citet{gupta-etal-2017-entity} additionally train context and description embeddings jointly, but this mainly aims to improve 
the quality of the fixed entity embeddings rather than 
using the context and description embeddings directly; 
we find that their context and description encoders 
perform poorly on \enteval tasks.

A closely related concurrent work by \citep{logeswaran-etal-2019-zero} jointly encodes 
a mention in context and an entity description 
from Wikipedia to perform zero-shot entity linking. 
In contrast, here we seek to pretrain a general purpose entity representations that can function well either 
given or not given entity descriptions or mention contexts.

Other entity-related tasks involve entity typing~\citep{yaghoobzadeh-schutze-2015-corpus,murtyfiner,delfinet,rabinovich2017fine,choi-etal-2018-ultra,onoe2019learning,obeidat-etal-2019-description} and coreference resolution~\citep{durrett-klein-2013-easy,wiseman-etal-2016-learning,lee-etal-2017-end,webster-etal-2018-mind,kantor-globerson-2019-coreference}. 

\paragraph{Evaluating pretrained representations.} 

Recent work has sought to evaluate the knowledge acquired by pretrained language models~\cite[\emph{inter alia}]{shi-etal-2016-string,adi2016fine,belinkov-etal-2017-evaluating,peters-etal-2018-dissecting,conneau-etal-2018-cram,conneau2018senteval,wang-etal-2018-glue,liu-gardner-belinkov-peters-smith:2019:NAACL,mchen-discoeval-19}. In this work, we focus on evaluating their capabilities in modeling entities. 

Part of \enteval involves evaluating world knowledge about entities, relating them to fact checking~\cite{vlachos-riedel-2014-fact,wang-2017-liar,thorne2018fever,yin-roth-2018-twowingos,chen-etal-2019-seeing}, and commonsense learning~\cite{angeli-manning-2014-naturalli,bowman-etal-2015-large,li-etal-2016-commonsense,mihaylov-etal-2018-suit,zellers2018swag,trinh2018simple,talmor-etal-2019-commonsenseqa,zellers-etal-2019-hellaswag,sap2019socialiqa,rajani2019explain}. Another related line of work is to integrate entity-related knowledge into the training of language models~\cite{logan-etal-2019-baracks,zhang-etal-2019-ernie,sun2019ernie}.

\paragraph{Contextualized word representations.} 

Contextualized word representations and pretrained language representation models, such as ELMo~\citep{peters-etal-2018-deep} and BERT~\citep{devlin2018bert}, are powerful pretrained models that have been shown to be effective for a variety of downstream tasks such as text classification, sentence relation prediction, named entity recognition, and question answering. 
Recent work has sought to evaluate the knowledge acquired by such models~\citep{shi-etal-2016-string,adi2016fine,belinkov-etal-2017-evaluating,conneau-etal-2018-cram,conneau2018senteval,liu-gardner-belinkov-peters-smith:2019:NAACL}. 
In this work, we focus on evaluating their capabilities in modeling entities. 
\section{\enteval}

We are interested in two approaches: \contextrep (henceforth: CER) and \descrep (henceforth: DER), both encoding fixed-length vector representations for entities. 

The \contextrep encodes an entity based on the context it appears regardless of whether the entity is seen before. The motivation behind \contextrep is that we want an entity encoder that does not depend on entries in a knowledge base, but 
is capable of inferring knowledge about an entity from the context it appears.

As opposed to \contextrep, \descrep do rely on entries in Wikipedia. We use a model-specific function $f$ to obtain a fixed-length vector representation 
from the entity's textual description.

To evaluate CERs and DERs, we propose a wide range of entity related tasks. Since our purpose is for examining the learned entity representations, we only use a linear classifier and freeze the entity representations when performing the following tasks. Unless otherwise noted, when the task involves a pair of entities, the input to the classifier are the entity representations $x_1$ and $x_2$, concatenated with their element-wise product and absolute difference: $[x_1, x_2, x_1 \odot x_2, |x_1 - x_2|]$. This input format has been used in SentEval~\cite{conneau2018senteval}.

\begin{table*}[t]
    \small
    \centering
\begin{tabular}{c|cc|c|c|c|c|cc|c|c|c}
\toprule
& \multicolumn{2}{c|}{ CAP } & CERP & EFP & ET & KORE & \multicolumn{2}{c|}{ WikiSRS } & ERT & Rare & CoNLL \\
& same & next & & & & & Rel & Sim & & & \\
\midrule
\#train & 3982 & 3982 & 4000 & 10000 & 1998 & N/A & N/A & N/A & 3130 & 10000 & 18538 \\
\#valid & 3806 & 3828 & 4000 & 2000 & 1998 & N/A & N/A & N/A & 6260 & 4000 & 4790 \\
\#test & 3938 & 3850 & 4000 & 2000 & 1998 & $20 \times 20$ & 688 & 688 & 6260 & 4000 & 4481 \\
\#classes & \multicolumn{2}{c|}{ 2 } & 2 & 2 & 10331 & N/A & N/A & N/A & 626 & 4 & up to 30 \\
\bottomrule
\end{tabular}
\caption{Statistics of datasets used in \enteval tasks. CAP: coreference arc prediction, CERP: contexualized entity relationship prediction, EFP: entity factuality prediction, ET: entity typing, ESR: entity similarity and relatedness, ERT: entity relationship typing, NED: named entity disambiguation, Rare: rare entity prediction, CoNLL: CoNLL-YAGO named entity disambiguation.}
\vspace{-0.1in}
 \label{table:dataset}
\end{table*}%

The datasets used in \enteval tasks are summarized in table~\ref{table:dataset}. 
It shows the number of instances in train/valid/test split for each dataset, and the number of target classes if this is a classification task. 
We describe the proposed tasks in the following subsections.

\subsection{Entity Typing (ET)}
\begin{figure}[t]
    \centering
    \includegraphics[scale=0.5]{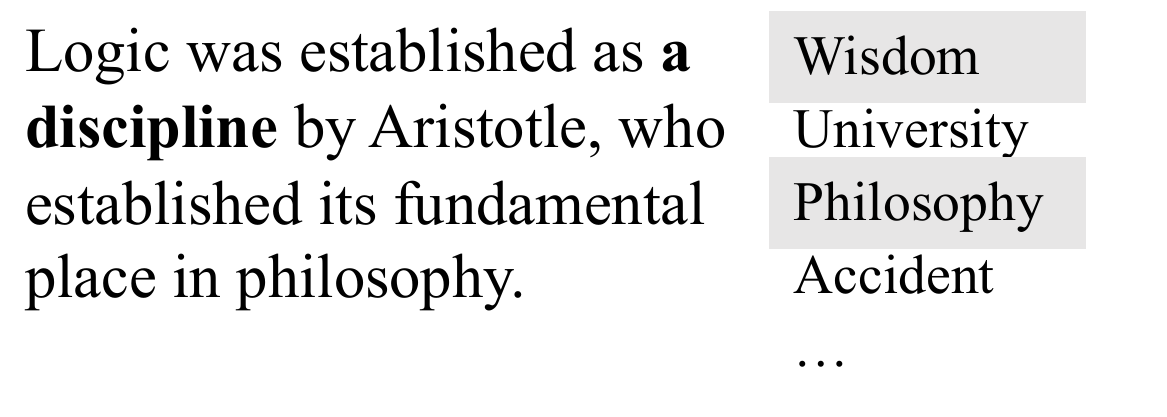}
    \caption{An example taken from ET. Targeted entity mention is bold. Candidate categories are on the right. Gold standard categories are in gray.}
    \vspace{-0.1in}
    \label{fig:entity-typing-example}
\end{figure}
The task of entity typing (ET) is to assign types to an entity given only the context of the entity mention. ET is context-sensitive, making it an effective approach to probe the knowledge of context encoded in pretrained representations. 
For example, in the sentence ``Bill Gates has donated billions to eradicate malaria'', 
``Bill Gates'' has the type of ``philanthropist'' instead of ``inventor''~\cite{choi-etal-2018-ultra}.

In this task, we will \contextrep, followed by a linear layer to make predictions. We use the annotated ultra-fine entity typing dataset of \citet{choi-etal-2018-ultra} with standard data splits. As shown in Figure~\ref{fig:entity-typing-example}, there can be multiple labels for an instance. We use binary log loss for training
using all positive and negative entity types, and report $F_1$ score. 
Thresholds are tuned based on validation set accuracy.

\subsection{Coreference Arc Prediction (CAP)}
Given two entities and the associated context, the task is to determine whether they refer to the same entity. Solving this task may require the knowledge of entities. For example, in the sentence ``Revenues of \$14.5 billion were posted by Dell\textsubscript{1}. The company\textsubscript{1} ...'', there is no prior context of ``Dell'', so having known ``Dell'' is a company instead of the people ``Michael Dell'' will surely benefit the model~\cite{durrett-klein-2014-joint}. Unlike other tasks, coreference typically involves longer context. To restrict the effect of broad context, 
we only keep two groups of coreference arcs from smaller context. One includes mentions that are in the same sentence (``same'') for examining the model capability of encoding local context. The other includes mentions that are in consecutive sentences (``next'') for the broader context. 
We create this task from the PreCo dataset~\cite{chen-etal-2018-preco}, which has mentions annotated even when they are not part of coreference chains. We filter out instances in which both mentions are pronouns. All non-coreferent mention pairs are considered to be negative samples.

To make this task more challenging, for each instance we compute cosine similarity of mentions by averaging GloVe word vectors. 
We group the instances into bins by cosine similarity, and 
randomly select the same number of positive and negative instances from each bin to ensure that models do not solve this task by simply comparing similarity of mention names.

We use the \contextrep of the two mentions to infer coreference arcs with supervised training and report the averaged accuracy of ``same'' and ``next''.

\subsection{Entity Factuality Prediction (EFP)}
The entity factuality prediction (EFP) task involves determining the correctness of statements regarding entities. 
We use the manually-annotated \fever dataset~\cite{thorne2018fever} for this task. 
\fever is a task to verify whether a statement is supported by evidences. 
The original \fever dataset includes three classes, namely ``Supports'', ``Refutes'', and ``NotEnoughInfo'' and evidences are additionally available for each instance. 
As our purpose is to examine the knowledge encoded in entity representations, we discard the last category (``NotEnoughInfo'') and the evidence. 
In rare cases, instances in \fever may include multiple entity mentions, so we randomly pick one. We randomly sample 10000, 2000, and 2000 instances for our training, validation, and test sets, respectively.

In this task, entity representations can be obtained either by \contextrep or \descrep. In practice, we observe \descrep give better performance, which presumably is because these statements are more similar to descriptions than entity mentions. 
As shown in Figure~\ref{fig:example-fever}, without providing additional evidences, solving this task requires knowledge of entities encoded in representations. We directly use entity representations as input to the classifier.

\begin{figure}[t]
    \small
    \centering
    \begin{Frame}
        \textit{REFUTES}: The \textcolor{violet}{\textbf{New York City Landmarks Preservation Commission}} consists of zero commissioners.\\
        \textit{SUPPORTS}: \textcolor{violet}{\textbf{TD Garden}} has held Bruins games.
    \end{Frame}
    \caption{Two examples from the EFP.}
    \label{fig:example-fever}
\end{figure}

\begin{figure}[t]
    \small
    \centering
    \begin{Frame}
    \textit{TRUE}: \textcolor{violet}{\textbf{Gin and vermouth}} can \textcolor{violet}{\textbf{make a martini}}\\
    \textit{FALSE}: \textcolor{violet}{\textbf{Connecticut}} is not \textcolor{violet}{\textbf{a state}}
    \end{Frame}
    \caption{Examples from the CERP.}
    \vspace{-0.05in}
    \label{fig:example-conceptnet}
\end{figure}

\subsection{Contexualized Entity Relationship Prediction (CERP)}

The task of contexualized entity relationship prediction (CERP) modeling determines the connection between two entities appeared in the same context.

We use sentences from \conceptnet~\cite{speer2017conceptnet} with automatically parsed mentions and templates used to construct the dataset. We filter out non-English concepts and relations such as `related', `translation', `synonym', and `likely to find' since we seek to evaluate more complicated knowledge of entities encoded in representations. We further filter out non-entity mentions and entities with type `DATE', `TIME', `PERCENT', `MONEY', `QUANTITY', `ORDINAL', and `CARDINAL' according to SpaCy~\cite{spacy2}. After filtering, we have 13374 assertions.

Negative samples are generated based on the following rules: 
\begin{enumeratesquish}
    \item For each relationship, we replace an entity with similar negative entities based on cosine similarity of averaged GloVe embeddings~\cite{pennington2014glove}.
    \item We change the relationship in positive samples from affirmation to negation (e.g., `is' to `is not'). These serve as negative samples. 
    \item We further sample positive samples from (1) in an attempt to prevent the `not' token from being biased towards negative samples. Therefore, for negative samples we get from (1), we change the relationship from affirmation to negation as in (2) to get positive samples.
\end{enumeratesquish}
    For example, let `A is B' be the positive sample. (1) changes it to `C is B' which serves as a negative sample and (2) changes it to `A is not B' as another negative sample. (3) changes it to 'C is not B' as a positive example.
In the end, we randomly sample 6000 instances from each class. This ends up yielding a 4000/4000/4000 train/dev/test dataset. As shown in Figure~\ref{fig:example-conceptnet}, this task cannot be solved by relying on surface form of sentences, instead it requires the input representations to encode knowledge of entities based on the context. 

We use \contextrep in this task.

\subsection{Entity Similarity and Relatedness (ESR)}
\begin{table}[t]
    \small
    \centering
    \begin{tabular}{c|c}
    \toprule
        Score & Entity Name \\
        \midrule
        - & Apple Inc. \\
        20 & Steve Jobs \\
        ... & ... \\
        11 & Microsoft \\
        ... & ... \\
        1 & Ford Motor Company\\
        \bottomrule
    \end{tabular}
    \caption{An example from \kore. The task is to rank the candidate entities by similarity.}
    \vspace{-0.1in}
    \label{tab:exmpale-kore}
\end{table}

Given two entities with their descriptions from Wikipedia, the task is to determine their similarity or relatedness. After the entity descriptions are encoded into vector representations, we compute their cosine similarity as predictions. 
We use the \kore~\citep{hoffart2012kore} and \wikisrs~\citep{newman-griffis-etal-2018-jointly} datasets in this task. Since the original datasets only provide entity names, we automatically add Wikipedia descriptions to each entity and manually ensure that every entity is matched to a Wikipedia description. 
We use Spearman's rank correlation coefficient between our computed cosine similarity and the gold standard similarity/relatedness scores to measure the performance of entity representations.

As \kore does not provide similarity scores of entity pairs, but simply ranks the candidate entities by their similarities to a target entity, we assign scores from 20 to 1 accordingly to each entity in the order of similarity. Table~\ref{tab:exmpale-kore} shows an example from \kore. The fact that ``Apple Inc.'' is more related to ``Steve Jobs'' than ``Microsoft'' requires multiple steps of inference, which motivates this task. Since the predictor we use is cosine similarity, which does not introduce additional parameters, we directly use encoded representations on the test set without any supervised training.

\subsection{Entity Relationship Typing (ERT)}
As another popular resource for common knowledge, we consider using Freebase~\citep{bollacker2008freebase} for probing the encoded knowledge by classifying the types of relations between pair of entities. First, we extract entity relation tuples (entity1, relation, entity2) from Freebase and then filter out easy tuples based on training a classifier using averaged GloVe vectors of entity names as input, which leaves us 626 types of relations, including ``internet.website.owner'', ``film.film\_art\_director.films\_art\_directed'', and ``comic\_books.comic\_book\_series.genre''. We randomly sample 5 instances for each relation type to form our training set and 10 instances per type the for validation and test sets. 
We use Wikipedia descriptions for each entity in the pair whose relation we are predicting and we use \descrep for each entity with supervised training.

\subsection{Named Entity Disambiguation (NED)}
Named entity disambiguation is the task of linking a named-entity mention to its corresponding instance in a knowledge base such as Wikipedia. In this task, we consider \yago~(CoNLL; \citealp{hoffart2011robust}) and Rare Entity Prediction~(Rare; \citealp{long2017world}).

\begin{figure}
    \centering
    \includegraphics[scale=0.4]{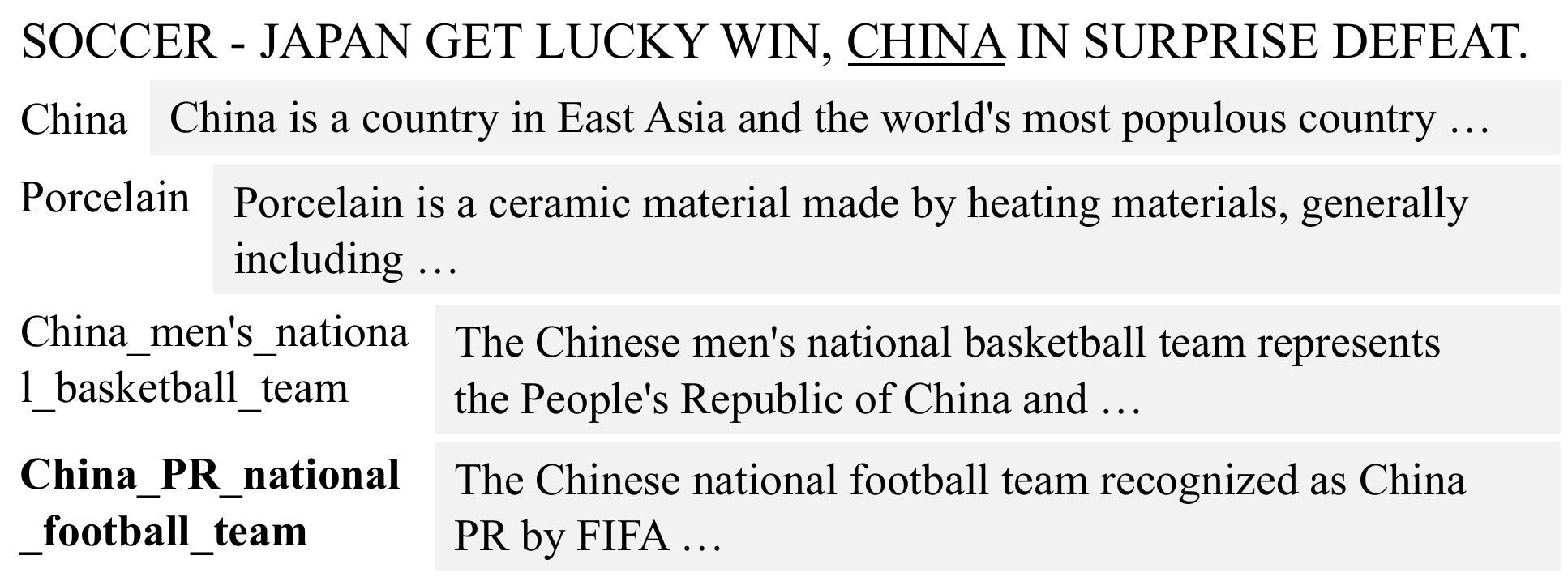}
    \caption{An example from \yago. Only four candidates are shown due to space constraints. The target mention is underlined. Sentences in gray are Wikipedia descriptions. The gold standard is boldfaced.}
    \vspace{-0.1in}
    \label{fig:conll-yago-example}
\end{figure}

For \yago, following \citet{hoffart2011robust} and \citet{yamada-etal-2016-joint}, we used the 27,816 mentions with valid entries in the knowledge base. For each entity mention $\mathit{m}$ in its context, we generate a set of (at most) its top 30 candidate entities $\mathit{C_m} = \{\mathit{c_j}\}$ using CrossWikis~\citep{spitkovsky-chang-2012-cross}. Some gold standard candidates $c$ are not present in CrossWikis, so we set the prior probability $p_\text{prior}(y)$ for those to 1e-6 and normalize the resulting priors for the candidate entities. When adding Wikipedia descriptions, we manually ensure gold standard mentions are attached to a description, however, we discard candidate mentions that cannot be aligned to a Wikipedia page. We use \contextrep for entity mentions and use \descrep for candidate entities. Training minimizes binary log loss 
using all negative examples. At test time, we use $\argmax_{c\in\mathit{C_m}}[p_\text{prior}(c)+p_\text{classifier}(c)]$ as the prediction. We note that directly using prior as predictions yields an accuracy of 58.2\%.

\citet{long2017world} introduce the task of \emph{rare entity prediction}. The task has a similar format to \yago entity linking. Given a document with a blank in it, the task is to select an entity from a provided list of entities with descriptions. Only rare entities are used in this dataset so that performing well on the task requires the ability to effectively represent entity descriptions. We randomly select 10k/4k/4k examples to construct train/valid/test sets. For simplicity, we only keep instances with four candidate entities. 

Figure~\ref{fig:conll-yago-example} shows an example from \yago, where 
the ``China'' in context has many deceptive meanings. Here the candidate ``China'' has exact string match of the entity name but it should not be selected as it is an after-game report on soccer. 
To match the entities, this task requires both effective contextualize entity representations and descriptive entity representation.

Practically, we encode the context using CER to be $x_1$, and encode each entity description using DER to be $x_2$, and pass $[x_1, x_2, x_1 \odot x_2, |x_1 - x_2|]$ to a linear model to predict whether it is the correct entity to fill in. 
The model is trained with cross entropy loss. 

\section{Methods}

We first describe how we define encoders for \contextrep (Section~\ref{sec:ctx-ent-rep}) and \descrep (Section~\ref{sec:desc-ent-rep}), then we discuss how we train new encoders tailored to capture information from the hyperlink structure of Wikipedia (Section~\ref{sec:hyperlink}). 

\subsection{Encoders for Contextualized Entity Representations}
\label{sec:ctx-ent-rep}
For defining these encoders, we assume we have a sentence $s = (w_1, \dots, w_T)$ where span $(w_i, \dots, w_j)$ refers to an entity mention. When using \elmo, we first encode the sentence: $(c_1, \dots, c_T) = \mathrm{ELMo}(w_1, \cdots, w_T)$, and we use the average of contextualized hidden states corresponding to the entity span as the contextualized entity representation. That is, $f_\text{\elmo}(w_{1:T},i,j)=\frac{\sum_{k=i}^j c_k}{j-i+1}$. 

With \bert, following \citet{onoe2019learning}, we concatenate the full sentence with the entity mention, starting with $\mathrm{[CLS]}$ and separating the two by $\mathrm{[SEP]}$, i.e.,  $\mathrm{[CLS]}, w_1, \dots, w_T, \mathrm{[SEP]}, w_i, \dots, w_j, \mathrm{[SEP]}$. We encode the full sequence using BERT and use the output from the $\mathrm{[CLS]}$ token as the entity mention  representation. 

\subsection{Encoders for Descriptive Entity Representations}
\label{sec:desc-ent-rep}
We encode an entity description by treating the entity description as a sentence, and use the average of the hidden states from \elmo as the entity description representation. 
With \bert, we use the output from the $\mathrm{[CLS]}$ token as the description representation. 

\subsection{Hyperlink-Based Training}
\label{sec:hyperlink}

An entity mentioned in a Wikipedia article is often linked to its Wikipedia page, which provides a useful description of the mentioned entity. 
\begin{figure}
    \centering
    \includegraphics[scale=0.45]{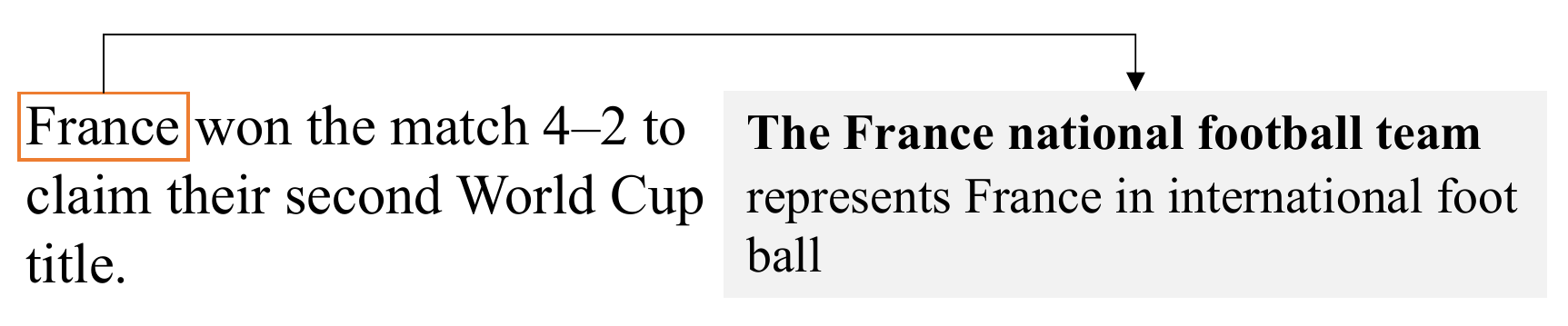}
    \caption{An example of hyperlinks in Wikipedia. ``France'' is linked to the Wikipedia page of ``France national football team'' instead of the country France.}
    \vspace{-0.1in}
    \label{fig:wiki-link}
\end{figure}
The same Wikipedia page may correspond to many different entity mentions. Likewise, the same entity mention may refer to different Wikipedia pages depending on its context. 
For instance, as shown in Figure~\ref{fig:wiki-link}, based on the context, ``France'' is linked to the Wikipedia page of ``France national football team'' instead of the country. 
The specific entity in the knowledge base can be inferred from the context information. 
In such cases, we believe Wikipedia provides valuable complementary information to the current pretrained CWRs such as \bert and \elmo.

To incorporate such information during training, we automatically construct a hyperlink-enriched dataset from Wikipedia that we will refer to as \wikilink. Prior work has used similar resources~\cite{singh12:wiki-links,gupta-etal-2017-entity}, but we aim to standardize the process and will release the dataset.

The \wikilink dataset consists of sentences with contextualized entity mentions and their corresponding descriptions obtained via hyperlinked Wikipedia pages. 
When processing descriptions, we only keep the first 100 word tokens at most as the description of a Wikipedia page; similar truncation has been done in prior work~\cite{gupta-etal-2017-entity}. 
For context sentences, we remove those without hyperlinks from the training data and duplicate those with multiple hyperlinks. We also remove context sentences for which we cannot find matched Wikipedia descriptions. These processing steps result in a training set of approximately 85 million instances and over 3 million unique entities. 

We define a hyperlink-based training objective and add it to \elmo. In particular, we use \contextrep to decode the hyperlinked Wikipedia description, and also use the \descrep to decode the linked context. We use bag-of-words decoders in both decoding processes. More specifically, given a context sentence $x_{1:T_x}$ with mention span $(i,j)$ and a description sentence $y_{1:T_y}$, we use the same bidirectional language modeling loss $l_\text{lang}(x_{1:T_x})+l_\text{lang}(y_{1:T_y})$ in \elmo where
\begin{align*}
    l_\text{lang}(u_{1:T})=-\sum_{t = 1}^T &\log p(u_{t+1}\vert u_1,\dots,u_{t}) + \\
    &\log p(u_{t-1}\vert u_{t},\dots,u_{T})    
\end{align*}

and $p$ is defined by the \elmo parameters. 
In addition, we define the two bag-of-words reconstruction losses:
\begin{align*}
    l_\text{ctx}  &= -\sum_t \log q(x_t\vert f_\text{\elmo}(\mathrm{[BOD]} y_{1:T_y}, 1, T_y)) \\
    l_\text{desc} &= -\sum_t \log q(y_t\vert f_\text{\elmo}(\mathrm{[BOC]} x_{1:T_x}, i, j))
\end{align*}
where $\mathrm{[BOD]}$ and $\mathrm{[BOC]}$ are special symbols prepended to sentences to distinguish descriptions from contexts.  
The distribution $q$ is parameterized by a linear layer that transforms 
the conditioning embedding into weights over the vocabulary. 
The final training loss is
\begin{align}
    l_\text{lang}(x_{1:T_x})+l_\text{lang}(y_{1:T_y})+l_\text{ctx}+l_\text{desc} \label{eq:loss}
\end{align}
Same as the original \elmo, each log loss is approximated with negative sampling~\cite{jean-etal-2015-using}. 
We write \entelmo to denote the model trained by Eq.~\eqref{eq:loss}. When using \entelmo for \contextrep  and \descrep, we use it analogously to \elmo. 

\section{Experiments}

\begin{table*}[t]
\small
    \centering\begin{tabular}{l|ccccccc|c}
    \toprule
& CAP & CERP & EFP & ET & ESR & ERT & NED & Average \\
\midrule
GloVe & 71.9 & 52.6 & 67.0 & 10.3 & 50.9 & 40.8 & 41.2 & 47.8 \\
BERT Base & \bf 80.6 & 65.6 & 74.8 & 32.0 & 28.8 & 42.2 & 50.6 & 53.5 \\
BERT Large & 79.1 & \bf 66.9 & \bf 76.7 & 32.3 & 32.6 & \bf 48.8 & \bf 54.3 & 55.8 \\
ELMo & 80.2 & 61.2 & 75.8 & \bf 35.6 & 60.3 & 46.8 & 51.6 & \bf 58.8 \\
\midrule
EntELMo baseline & 78.0 & 59.6 & 71.5 & 31.3 & \bf 61.6 & 46.5 & 48.5 & 56.7 \\
\entelmo & 76.9 & 59.9 & 72.4 & 32.2 & 59.7 & 45.7 & 49.0 & 56.5 \\
\entelmo w/o $l_\text{ctx}$ & 73.5 & 59.4 & 71.1 & 33.2 & 53.3 & 44.6 & 48.9 & 54.9 \\
\entelmo w/ $l_\text{etn}$ & 76.2 & 60.4 & 70.9 & 33.6 & 49.0 & 42.9 & 49.3 & 54.6 \\
\bottomrule
\end{tabular}
    \caption{Performances of entity representations on \enteval tasks. Best performing model in each task is boldfaced. CAP: coreference arc prediction, CERP: contexualized entity relationship prediction, EFP: entity factuality prediction, ET: entity typing, ESR: entity similarity and relatedness, ERT: entity relationship typing, NED: named entity disambiguation. 
    \entelmo baseline is trained on the same dataset as \entelmo but not using the hyperlink-based training. \entelmo w/ $l_\text{etn}$ is trained with a modified version of $l_\text{ctx}$, where we only decode entity mentions instead of the whole context.
    }
    \vspace{-0.1in}
    \label{tab:results}
\end{table*}

\subsection{Setup}

As a baseline for hyperlink-based training, we train \entelmo on the \wikilink dataset with only a bidirectional language model loss. Due to the limitation of computational resources, both variants of \entelmo are trained for one epoch (3 weeks time) with smaller dimensions than \elmo. We set the hidden dimension of each directional long short-term memory network~(LSTM; \citealp{hochreiter1997long}) layer to be 600, and project it to 300 dimensions.
The resulting vectors from each layer are thus 600 dimensional. We use 1024 as the negative sampling size for each positive word token. For bag-of-words reconstruction, we randomly sample at most 50 word tokens as positive samples from the the target word tokens. Other hyperparameters are the same as \elmo. \entelmo is implemented based on the official \elmo implementation.\footnote{Our implementation is available at \url{https://github.com/mingdachen/bilm-tf}}

As a baseline for contextualized and descriptive entity representations, we use GloVe word averaging of the entity mention as the ``contextualized'' entity representation, and use word averaging of the truncated entity description text as its description representation. We also experiment two variants of \entelmo, namely \entelmo w/o $l_\text{ctx}$ and \entelmo with $l_\text{etn}$. For second variant, we replace $l_\text{ctx}$ with $l_\text{etn}$, where we only decode entity mentions instead of the whole context from descriptions.
We lowercased all training data as well as the evaluation benchmarks.

We evaluate the transferrability of \elmo, \entelmo, and \bert by using trainable mixing weights for each layer. For \elmo and \entelmo, we follow the recommendation from \citet{peters-etal-2018-deep} to first pass mixing weights through a softmax layer and then multiply the weighted-summed representations by a scalar. For \bert, we find it better to just use unnormalized mixing weights. In addition, we investigate per-layer performance for both models in Section~\ref{sec:analysis}.

\subsection{Results}

Table~\ref{tab:results} shows the performance of our models on the \enteval tasks. Our findings are detailed below:
\begin{itemizesquish}
\item Pretrained CWRs (ELMo, BERT) perform the best on \enteval overall, indicating that they capture knowledge about entities in contextual mentions or as entity descriptions. 
\item BERT performs poorly on entity similarity and relatedness tasks. 
Since this task is zero-shot, it validates the recommended setting of finetuning BERT~\citep{devlin2018bert} on downstream tasks, while the embedding of the $\mathrm{[CLS]}$ token does not necessarily capture the semantics of the entity.
\item BERT Large is better than BERT Base on average, showing large improvements in ERT and NED. To perform well at ERT, a model must either glean particular relationships from pairs of lengthy entity descriptions or else leverage knowledge from pretraining about the entities considered.  Relatedly, performance on NED is expected to increase with both the ability to extract knowledge from descriptions and by starting with increased knowledge from pretraining. The Large model appears to be handling these capabilities better than the Base model.

\item \entelmo improves over the \entelmo baseline (trained without the hyperlinking loss) on some tasks but suffers on others. The hyperlink-based training helps on CERP, EFP, ET, and NED. Since the hyperlink loss is closely-associated to the NED problem, it is unsurprising that NED performance is improved. 
Overall, we believe that hyperlink-based training benefits contextualized entity representations but does not benefit descriptive entity representations (see, for example, the drop of nearly 2 points on ESR, which is based solely on descriptive representations). This pattern may be due to the difficulty of using descriptive entity representations to reconstruct their appearing context.

\end{itemizesquish}

\section{Analysis}\label{sec:analysis}

\paragraph{Is descriptive entity representation necessary?}

\begin{table}[t]
\centering
\setlength{\tabcolsep}{4pt}
\small
\begin{tabular}{c|cc|cc|cc}
\toprule
& \multicolumn{2}{c|}{ Rare } & \multicolumn{2}{c|}{CoNLL} & \multicolumn{2}{c}{ERT}\\
& Des. & Name & Des. & Name & Des. & Name \\
\midrule
\elmo & 38.1 & 36.7 & 63.4 & 71.2 & 46.8 & 31.5\\
\bert Base & 42.2 & 36.6 & 64.7 & 74.3 & 42.2 & 34.3\\
\bert Large & 48.8 & 44.0 & 64.6 & 74.8 & 48.8 & 32.6 \\ 
\bottomrule
\end{tabular}
\caption{Accuracies (\%) in comparing the use of description encoder (Des.) to entity name (Name).}
    \label{tab:name_vs_desc}
\end{table}
A natural question to ask is whether the entity description is needed, as for humans, the entity names carry sufficient amount of information for a lot of tasks. 
To answer this question, we experiment with encoding entity names by the descriptive entity encoder for ERT (entity relationship typing) and NED (named entity disambiguation) tasks. The results in Table~\ref{tab:name_vs_desc} show that encoding the entity names by themselves already captures a great deal of knowledge regarding entities, especially for \yago. However, in tasks like ERT, the entity descriptions are crucial as the names do not reveal enough information to categorize their relationships.

\begin{table}[t]
\centering
\setlength{\tabcolsep}{5pt}
\small
\begin{tabular}{c|c}
\toprule
& CoNLL\\
\midrule
\elmo & 71.2 \\
\citet{gupta-etal-2017-entity} & 65.1 \\
Deep ED & 66.7 \\
\bottomrule
\end{tabular}
\caption{Accuracies (\%) on \yago with static or non-static entity representations.}
    \vspace{-0.1in}
    \label{tab:conll_yago}
\end{table}

Table~\ref{tab:conll_yago} reports the performance of different descriptive entity representations on the \yago task. 
The three models all use ELMo as the context encoder. 
``ELMo'' encodes the entity name with ELMo as descriptive encoder, while both \citet{gupta-etal-2017-entity} and Deep ED~\citep{ganea-hofmann-2017-deep} use their trained static entity embeddings. \footnote{We note that the numbers reported here are not strictly comparable to the ones in their original paper since we keep all the top 30 candidates from Crosswiki while prior work employs different pruning heuristics. }
As \citet{gupta-etal-2017-entity} and Deep ED have different embedding sizes from ELMo, 
we add an extra linear layer after them to map to the same dimension. These two models are designed for entity linking, which gives them potential advantages. Even so, ELMo outperforms them both by a wide margin.

\paragraph{Per-Layer Analysis.}
We evaluate each \elmo and \entelmo layer, i.e., the character CNN layer and two bidirectional LSTM layers, as well as each \bert layer on the \enteval tasks.
\begin{figure}[t]
    \centering
    \includegraphics[scale=0.4]{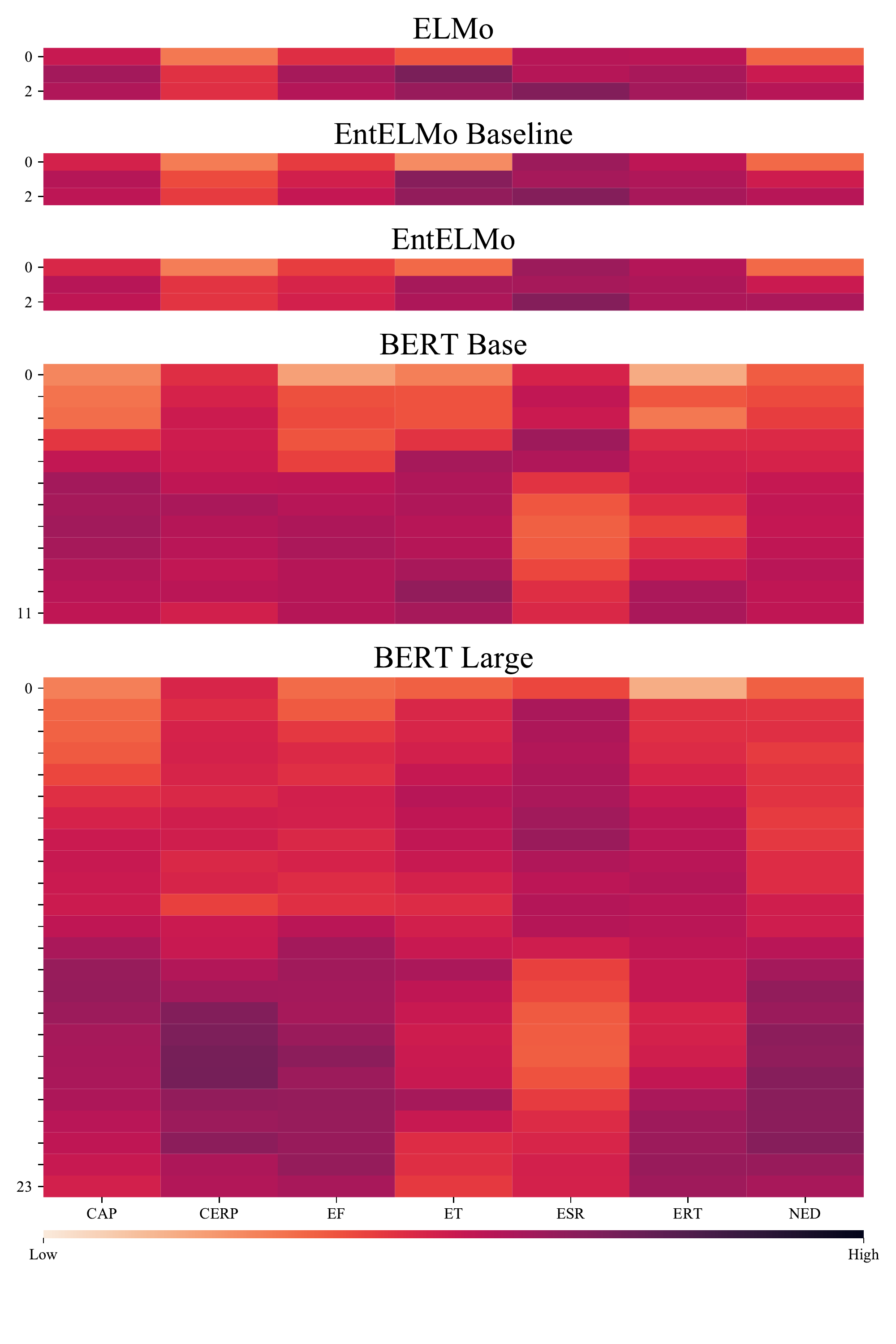}
    \vspace{-0.4in}
    \caption{Heatmap showing per-layer performances for \elmo, \entelmo baseline, \entelmo, \bert Base, and \bert Large. }
    \vspace{-0.1in}
    \label{fig:per-layer-analysis}
\end{figure}
Figure~\ref{fig:per-layer-analysis} reveals that for ELMo models, the first and second LSTM layers capture most of the entity knowledge from context and descriptions. 
The BERT layers show more diversity. Lower layers perform better on ESR (entity similarity and relatedness), while for other tasks higher layers are more effective. 

\section{Conclusion}
Our proposed \enteval test suite provides a standardized evaluation method for entity representations. 
We demonstrate that \enteval tasks can benefit from the success of contextualized word representations such as \elmo and \bert. 
Augmenting encoding-decoding loss leveraging natural hyperlinks from Wikipedia further improves \elmo on some \enteval tasks. 
As shown by our experimental results, the contextualized entity encoder benefits more from this hyperlink-based training objective, suggesting future works to prioritize encoding entity description from its mention context.

\section*{Acknowledgments}

We thank Davis Yoshida for discussions at the early stages of this project, 
Kushal Arora and Jackie Chi Kit Cheung for answering our questions on the Rare Entity Prediction dataset, and Nitish Gupta for clarifying details about models from \citet{gupta-etal-2017-entity}. This research was supported in part by a Bloomberg data science research grant to K.~Stratos and K.~Gimpel. 

\bibliography{emnlp-ijcnlp-2019}
\bibliographystyle{acl_natbib}

\end{document}